\documentclass[preprint,12pt]{elsarticle}

\usepackage{lineno,hyperref}
\usepackage{natbib}
\usepackage{geometry}
\usepackage{fleqn}
\usepackage{graphicx}
\usepackage{newtxtext,newtxmath}
\usepackage{hyperref}
\usepackage{booktabs}
\usepackage{bm}
\usepackage{amsmath}
\usepackage{enumitem}
\usepackage{esvect}
\usepackage{multicol}
\usepackage{multirow}
\usepackage{soul,xcolor}
\usepackage[justification=centering]{caption}
\usepackage{cleveref}
\modulolinenumbers[5]

\graphicspath{{./figures/}}
\journal{The International Journal of Advanced Manufacturing Technology}

\begin{document}

\begin{frontmatter}

\title{\large{Fast and Accurate Reduced-Order Modeling of a MOOSE-based Additive Manufacturing Model with Operator Learning}}

\author[NCSU]{Mahmoud Yaseen}
\author[INL1]{Dewen Yushu}
\author[INL2]{Peter German}
\author[NCSU]{Xu Wu\corref{mycorrespondingauthor}}
\cortext[mycorrespondingauthor]{Corresponding author}
\ead{xwu27@ncsu.edu}

\address[NCSU]{Department of Nuclear Engineering, North Carolina State University    \\ 
	Burlington Engineering Laboratories, 2500 Stinson Drive, Raleigh, NC 27695 \\}

\address[INL1]{Computational Mechanics \& Materials Department \\
	Idaho National Laboratory, P.O. Box 1625, Idaho Falls, ID 83415}

\address[INL2]{Computational Frameworks Department \\
	Idaho National Laboratory, P.O. Box 1625, Idaho Falls, ID 83415}

\begin{abstract}
One predominant challenge in additive manufacturing (AM) is to achieve specific material properties by manipulating manufacturing process parameters during the runtime. Such manipulation tends to increase the computational load imposed on existing simulation tools employed in AM. The goal of the present work is to construct a fast and accurate reduced-order model (ROM) for an AM model developed within the Multiphysics Object-Oriented Simulation Environment (MOOSE) framework, ultimately reducing the time/cost of AM control and optimization processes. Our adoption of the operator learning (OL) approach enabled us to learn a family of differential equations produced by altering process variables in the laser's Gaussian point heat source. More specifically, we used the Fourier neural operator (FNO) and deep operator network (DeepONet) to develop ROMs for time-dependent responses. Furthermore, we benchmarked the performance of these OL methods against a conventional deep neural network (DNN)-based ROM. Ultimately, we found that OL methods offer comparable performance and, in terms of accuracy and generalizability, even outperform DNN at predicting scalar model responses. The DNN-based ROM afforded the fastest training time. Furthermore, all the ROMs were faster than the original MOOSE model yet still provided accurate predictions. FNO had a smaller mean prediction error than DeepONet, with a larger variance for time-dependent responses. Unlike DNN, both FNO and DeepONet were able to simulate time series data without the need for dimensionality reduction techniques. The present work can help facilitate the AM optimization process by enabling faster execution of simulation tools while still preserving evaluation accuracy.
\end{abstract}


\end{frontmatter}


\section{Introduction}
\label{section:Introduction}

Additive manufacturing (AM) is a broad concept that encompasses various production technologies usable to create material parts layer by layer. Such technologies include fused deposition modeling, stereolithography, direct energy deposition (DED), and powder bed fusion. AM processes allow for components to be manufactured out of a wide range of materials, including metals, polymers, ceramics, and composites. This versatility has led to the emergence of AM applications in such diverse fields as medical surgery \cite{tack20163d}, hydrogen storage \cite{free2021review}, and aerospace engineering \cite{kestila2018towards}. Compared against subtractive methods such as milling and drilling, AM has revolutionized the traditional manufacturing paradigm by enabling fabrication of stronger and more efficient parts for complex systems, reducing operating times/costs, and minimizing waste.

Several complex physical processes interact during the AM process, yet our understanding of them remains limited, thus impeding the expansion of industrial-scale applications. These processes, which occur at the microstructural level, influence the thermal and mechanical properties, geometry shaping, and residual stress of the materials. The thermo-mechanical behavior of the deposited material is modeled by heat and stress partial differential equations (PDEs) typically solved via conventional numerical methods (e.g., finite element methods). As the manufacturing process involves making continuous changes to input and control parameters, optimizing the final product necessitates repetitive evaluation of the computational model. This in turn demands high-performance computing resources and prolonged simulation time/costs. To address this challenge, reduced-order models (ROMs) can be employed for multi-query tasks as substitutes for the original computational model (i.e., the full model). This makes control and optimization processes more time- and cost-efficient. The ROM development covered in the present work aims to achieve high-accuracy predictions, reduced execution times, and seamless integration into control and optimization processes.

Machine learning (ML), an artificial intelligence (AI) subset in which a machine learns from data in order to perform tasks such as making predictions based on a given input or engaging in decision-making activities, has been extensively applied when building ROMs in many fields, including the AM \cite{era2022prediction}, nuclear engineering \cite{yaseen2023quantification}, and fluid dynamics \cite{xiao2019reduced} fields. Generally, ML-based ROMs are more accurate than simple ROMs (e.g., polynomial-based ROMs), as ML models can be developed and customized in order to learn important non-linear patterns or extract certain desired features. For example, an autoencoder is used to extract important features by representing data on lower dimensional states, as was the case in \cite{zhao2022machine} for nonlinear systems. 

Recently, the U.S. Department of Energy and the nuclear industry have seen significant interest in applying AM approaches to the manufacture of nuclear materials (e.g., fuels, cladding, and structure materials) \cite{lou2019advanced} \cite{rodriguez2020metal} \cite{raftery2021fabrication}. In the present work, we focus on DED, a process also called laser-engineered net shaping, direct metal deposition, and laser consolidation. A novel, geometry-free thermo-mechanical model recently developed \cite{yushu2022directed} at Idaho National Laboratory (INL) uses adaptive subdomain construction to accurately predict the thermal conditions, distortions, and residual stresses that occur throughout the DED process. This model is based on the open-source Multiphysics Object-Oriented Simulation Environment (MOOSE) framework \cite{lindsay20222} developed at INL. The goal of the present work is to build an accurate, fast-running ROM for this high-fidelity but computationally prohibitive MOOSE-based AM model.

Integrating AM technology into the manufacture of nuclear materials poses certain challenges. One such challenge involves the need to continuously adjust control parameters throughout the manufacturing process in order to achieve the specific desired properties. Traditional hardware level controllers (e.g., proportional-integral-derivative controllers), which are designed to follow predetermined system settings, lack the ability to respond to dynamic process conditions. In contrast, intelligent AM aims to minimize human intervention in the optimization process, but relies on an automated process-level control mechanism to generate optimal design variables and adaptive system settings for improved end-product properties. An ongoing INL Laboratory Directed Research and Development project is currently working to develop a novel AI-based process control and optimization methodology that can be applied to AM processes, instead of implementing the existing trial-and-error approach. This AI-based process control and optimization methodology requires rapid evaluation of the expected AM results, and will thus benefit from the ROM, which requires much less computational time than the full model.

ML-based approaches have been developed to build efficient ROMs for computationally expensive models. These ML approaches can be broadly categorized as either ``physics-informed'' or ``data-driven''. The first type preserves the ``physics'' of the physical model---namely, by using physics-informed ML \cite{karniadakis2021physics} (e.g., physics-informed neural networks [PINNs]) \cite{raissi2019physics}. PINNs directly solve the differential equations while also enforcing the physics from the initial/boundary conditions and conservation equations in the loss function when training the deep neural networks (DNNs). However, PINNs work best for simple geometries and clearly defined differential equations. In the present application, the AM process will cause geometric changes, due to the formation of a melt pool. For the present work, other physics-informed-ML-related challenges include modeling the moving laser power source and tracking the melt pool surface, or interfacing with the substrate. It is thus difficult to implement the convective boundary condition. Moreover, PINNs can only solve one instance of the governing equations at a time. In the MOOSE-based AM model, the thermal model uses a Gaussian point heat source that has multiple process variables (e.g., laser power and laser scanning speed) to be controlled. Changing the values of these process variables produces a different set of equations on which the PINNs must be retrained. These issues motivated us to consider the second option, which is to build a data-driven ROM based on operator learning (OL). OL learns an operator, which can represent a family of differential equations. In other words, the process variables in the Gaussian point heat source are treated as inputs. When the governing PDEs change due to changes in these inputs, the OL-based ROM does not have to be retrained. Moreover, OL-based ROMs are data-driven, making them easier to train because they do not solve the differential equations directly, but instead use the input-output paired training data collected from physical model simulations. In particular, we employed the Fourier neural operator (FNO) and Deep Operator Network (DeepONet)---two OL methods that have been successfully demonstrated in other areas---while implementing certain modifications in light of our intended applications.

It must be noted that ML has been widely used in AM modeling and simulation problems. For example, in image processing, \cite{patel2019using} utilized convolutional neural networks (NNs) to analyze cross-sectional image data pertaining to the printed layers so as to identify regions containing dross. Another study \cite{khanzadeh2018porosity} applied many supervised learning algorithms to classify melt pool images as either normal or porous. Subsequently, the trained model can predict the porosity in a given part. \cite{schoinochoritis2017simulation} used a computer vision algorithm to identify and categorize anomalies that occur in the laser power bed fusion process. They used image patches as a training set for an unsupervised ML algorithm. Other studies used ML to predict product properties and AM process characteristics. In \cite{era2022prediction}, the authors employed two data-driven methodologies to predict the tensile behaviors (e.g., yield strength, ultimate tensile strength, and elongation [\%]) of stainless steel 316L parts manufactured via DED. \cite{barrionuevo2022machine} used a Gaussian process regressor to predict the melting efficiency of wire arc AM. A physics-based NN in \cite{zhu2021machine} was used to predict temperature and melt pool fluid dynamics. However, studies conducted on ML-based ROMs of expensive AM models remain relatively limited, and even less work has been applied to studying the applicability of OL-based ROMs for AM. Furthermore, in the present work, we implemented unique changes to the OL approaches in order to make them applicable for time-dependent AM quantities of interest (QoIs)---namely, bead volume and maximum melt pool temperature. Thus, we consider this work to be novel as well as important to the AM community.

For this paper, we utilized two OL methods in light of their capability to learn a family of differential equations by approximating the operator itself rather than learning on finite datasets. More specifically, we used modified FNO and DeepONet structures to develop ROMs for time-dependent responses. Furthermore, we compared the performance of these OL methods against a conventional DNN-based ROM used for making scalar response predictions. This work is an extension of our previous work, in which we applied FNO to the same MOOSE-based AM model and benchmarked it against the DNN-based ROM \cite{yaseen2023mooseam}. For conducting a performance evaluation of the two test cases, we compared the ROM results against each other in terms of root mean square error (RMSE), relative error, coefficient of determination ($R^2$), and time efficiency. These metrics indicate how well the ROMs captured the details in the original model, and also reflect the reduction in computational power. OL methods were found to offer comparable performance, and in terms of accuracy and generalizability, even outperformed the DNN at predicting scalar model responses. Moreover, all the ROMs were faster than the original MOOSE model yet still provided accurate predictions. FNO had a smaller mean prediction error than DeepONet, with a larger variance for time-dependent responses. And both FNO and DeepONet were able to simulate time series data without the need for dimensionality reduction techniques such as DNN. 

The remainder of this paper is organized as follows. Section \ref{section:MOOSE-AM-model} provides the theoretical background on the governing PDEs solved in the MOOSE-based AM model, and it describes the ROM inputs/outputs. Section \ref{section:Methodology} provides mathematical details regarding the OL and DNN approaches to building ROMs. Section \ref{section:Data-generation-and-pre-processing} provides details on collecting and preprocessing the training data. Section \ref{section:Results} presents the FNO and DeepONet results and compares them with the results obtained from regular DNNs, based on several different performance metrics. Section \ref{section:Conclusions} summarizes the findings covered in this paper.

\section{Description of the Computational MOOSE-based AM Model}
\label{section:MOOSE-AM-model}

The ROM development in the present work was based on the existing MOOSE-based AM model presented in \cite{yushu2022directed}.
This model is comprised of two key components: the thermal model and the mechanical model. The thermal model simulates laser-induced heating and cooling of the material by solving the heat conduction equation via a moving Gaussian heat source that mimics the laser scan. The mechanical model solves the quasi-static momentum conservation equation in order to compute the displacement and stress fields caused by thermal expansion. The thermal and mechanical models are coupled during the simulation process. Specifically, the material deformation and stress fields are dependent on the temperature change reflected in the thermal model, making this problem multiphysical in nature. At each time step, the heat conduction equation is solved to obtain the temperature field. The model then uses the subdomain construction approach \cite{yushu2022directed}, in which each element from the part being manufactured is switched from an inactive state to an active state if its average temperature is above the melt temperature. At the same time, adaptive mesh refinement is employed to ensure high solution accuracy at a low computational cost.

The corresponding thermal governing equation is written as follows:
\begin{align}    \label{eqn:heat-conduction}
        \rho c (T) &= \frac{\partial T }{\partial t} = \nabla \cdot k (T) \nabla T + Q (\bm{x}, t) \: \text{in} \: \Omega,\nonumber \\
        T &= \Bar{T} \: \text{on} \: \partial \Omega_{s, \text{bot}},\\
        - k (T) \nabla T \cdot \bm{n} &= h (T) \cdot \left(T - \Bar{T}_{\infty}\right) \: \text{on } \Gamma_{a, u} \nonumber
\end{align}
where $T$ is the temperature, $t$ is the time, $\bm{x} \in \mathbb{R}^{3}$ is the spatial location, $\bm{n} \in \mathbb{R}^{3}$ is the outward normal, $\rho$ is the material density, $c(T)$ is the temperature-dependent specific heat, $k (T) $ is the temperature-dependent thermal conductivity, and $h (T)$ is the temperature-dependent heat convection coefficient. At the bottom of the substrate ($\partial \Omega_{s, \text{bot}}$), the temperature is fixed at room temperature $\Bar{T}$. $\Bar{T}_{\infty}$ denotes the temperature far from the convective boundary, and $\Gamma_{a, u}$ represents the interface of the growing deposited material. $Q (\bm{x}, t) $ is a temporally and spatially varying heat source that mimics the scanning laser beam used in the DED process.

The heat source is a combination of the Gaussian point heat source model and the Gaussian line heat source model. The Gaussian point heat source is defined as:
\begin{equation}\label{eqn:Gaussian-point-heat-source}
  \hat{Q} (\bm{x},t)  =  \frac{2 \alpha \eta P}{\pi r^3} \exp \left(  - \frac{2 || \bm{x} - \bm{p}(t) ||^2}{r^2}  \right)
\end{equation}
where $P$ is the laser power, $\alpha$ is the equipment-related scaling factor, $\eta$ is the laser efficiency coefficient, $r$ is the effective radius of the laser beam, $|| \cdot||$ is the Euclidean norm, and $\bm{p}$ is the time-varying scanning path. Here, $\bm{p}$ is a time-dependent spatial vector that indicates the laser's scanning path, which depends on the part geometry and scanning velocity. In this study, we consider a constant scanning speed along the $y$-direction (i.e., $\bm{p}_y(t) = vt$, where $v$ is the scanning speed).

\begin{figure}[!htb]
	\centering
	\captionsetup{justification=centering}
	\includegraphics[width=1.0\textwidth] 
        {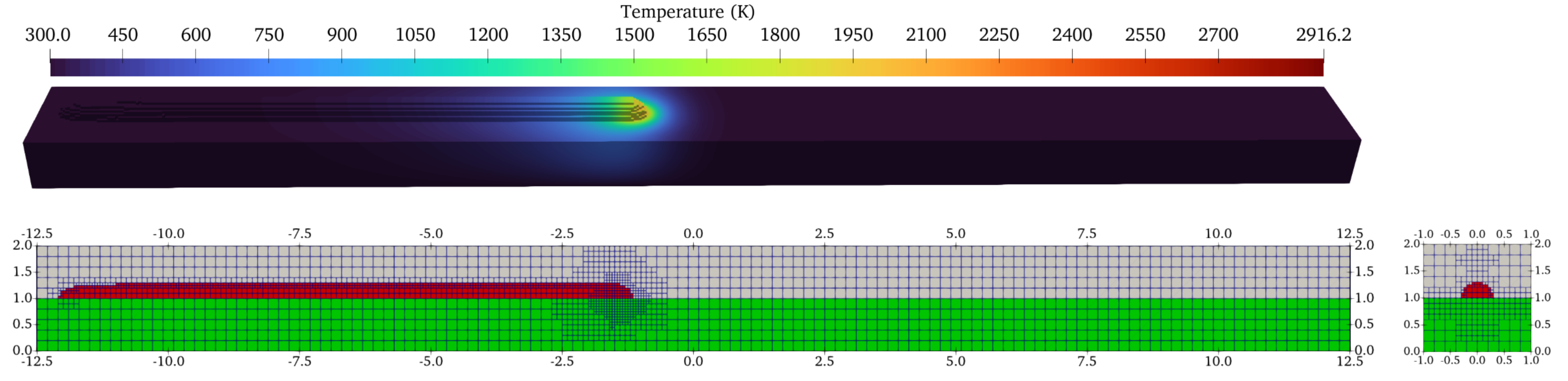}
	\caption[]{Visualization of the material printed via the DED 
         process, as simulated by the MOOSE-based AM model.}
	\label{fig:printed-material-visual}
\end{figure}

If time step $\Delta t$ is larger than that specified by the parameter values ($\Delta t > r/v$), the Gaussian point heat source defined in Equation \ref{eqn:Gaussian-point-heat-source} will skip over certain elements. To address this issue, the following Gaussian line heat source $\left(\Bar{Q}\left(\bm{x}, t\right) \right)$ was developed in \cite{irwin2016line}:
\begin{equation} \label{eqn:Gaussian-line-heat-source}
    \Bar{Q} (\bm{x}, t) = \frac{1}{\Delta t} \int_{t_{0}}^{t_{0} + \Delta t} \hat{Q} (\bm{x},t) \: dt
\end{equation}
where $t_{0}$ is the time at the beginning of the time step. In Equation  \ref{eqn:Gaussian-line-heat-source}, the line source is simply the average of the point source, such that $\lim_{\Delta t \to 0} \Bar{Q} (\bm{x}, t) = \hat{Q} (\bm{x},t)$. To avoid the numerical inaccuracy of replacing the point source with the line source, \cite{yushu2022directed} developed a hybrid heat source model that depends on the location of the laser point along the scanning path, switching between the line and point heat sources when needed. Figure \ref{fig:printed-material-visual} shows a visualization of the MOOSE simulation output, as generated by ParaView \cite{ayachit2015paraview}. The substrate's far end is at room temperature, and that temperature increases as the laser moves. Once the temperature of the powder material reaches the melting point, a melt pool forms, then solidifies once the laser moves away.

The present work focuses on the development of ROMs for the thermal model only (i.e., \Cref{eqn:heat-conduction,eqn:Gaussian-point-heat-source,eqn:Gaussian-line-heat-source}). There are three reasons for this. First, the thermal model plays a crucial role in modeling the material addition process, whereas the mechanical aspect in no way affects the material addition. Second, creating a universally useful mechanical model is impossible in light of the numerous forms of constitutive equations corresponding to different temporal and spatial scales, material types, and operating conditions. Moreover, creating a ROM hat specifically models the thermal  behavior enables us to retain the flexibility of coupling with various mechanical models (both high-fidelity ones and ROMs) \cite{hernandez2021model}. Finally, proper application of ROMs in multiphysics simulations is a challenge in and of itself, and is beyond the scope of creating an accurate and efficient ROM of the DED process.

To create a ROM that is applicable to different manufacturing conditions, five laser-heat-source-related parameters were chosen as inputs (see \Cref{eqn:Gaussian-point-heat-source,eqn:Gaussian-line-heat-source}): laser power ($P$), scanning speed ($v$), laser effective radius ($r$), laser efficiency coefficient ($\eta$), and scaling factor ($\alpha$). These parameters and their associated values are listed in Table \ref{table:AM-process-variables}. 

\begin{table}[htbp!]
        \footnotesize
	\centering
	\caption{Process parameters that will serve as inputs to the ROMs.}
	\label{table:AM-process-variables}
	\begin{tabular}{l c  c  c  c}
		\toprule
		Parameters & Symbols &  Nominal values & Bounds & Units\\
		\midrule
		Laser power  & $P$  & 300  & [250, 400] & W\\ 
		Scanning speed & $v$ & 0.01058 & [0.004, 0.020] & mm/ms\\
		Laser beam radius & $r$ & 0.3 &  [0.25, 0.40] & mm \\
		Laser efficiency coefficient & $\eta$ & 0.36  & [0.3, 0.4] & -\\
		Scaling factor  & $\alpha$ & 1.6 & [1.0, 2.0] & - \\
		\bottomrule
	\end{tabular}
\end{table}

To compare the effectiveness of different types of ROMs, we considered two different cases. \textbf{The first case} involved building ROMs for the maximum values of the QoIs over time. This is because the maximum bead volume (usually seen at the end of the simulation) and maximum melt pool temperature are the most relevant responses to consider when using the ROMs for AM process control and optimization. \textbf{The second case} involved building ROMs for the whole time series of the time-dependent QoIs. This is because the manner in which the two QoIs change over time is still of high interest in regard to designing algorithms for AM process control and optimization. However, a regular DNN-based ROM is inappropriate in this case, since for 200 time steps each QoI must be treated as 200 separate responses. (The response dimension could have been much higher had we chosen a larger number of time steps.) We modified the original DeepONet and FNO architectures to make them applicable for cases in which the responses are time dependent. Recall that most previous research on DeepONet and FNO has focused on spatially varying responses rather than temporally varying ones. 

\begin{figure}[!htb]
    \centering
    \captionsetup{justification=centering}
    \includegraphics[width=0.8\textwidth] 
    {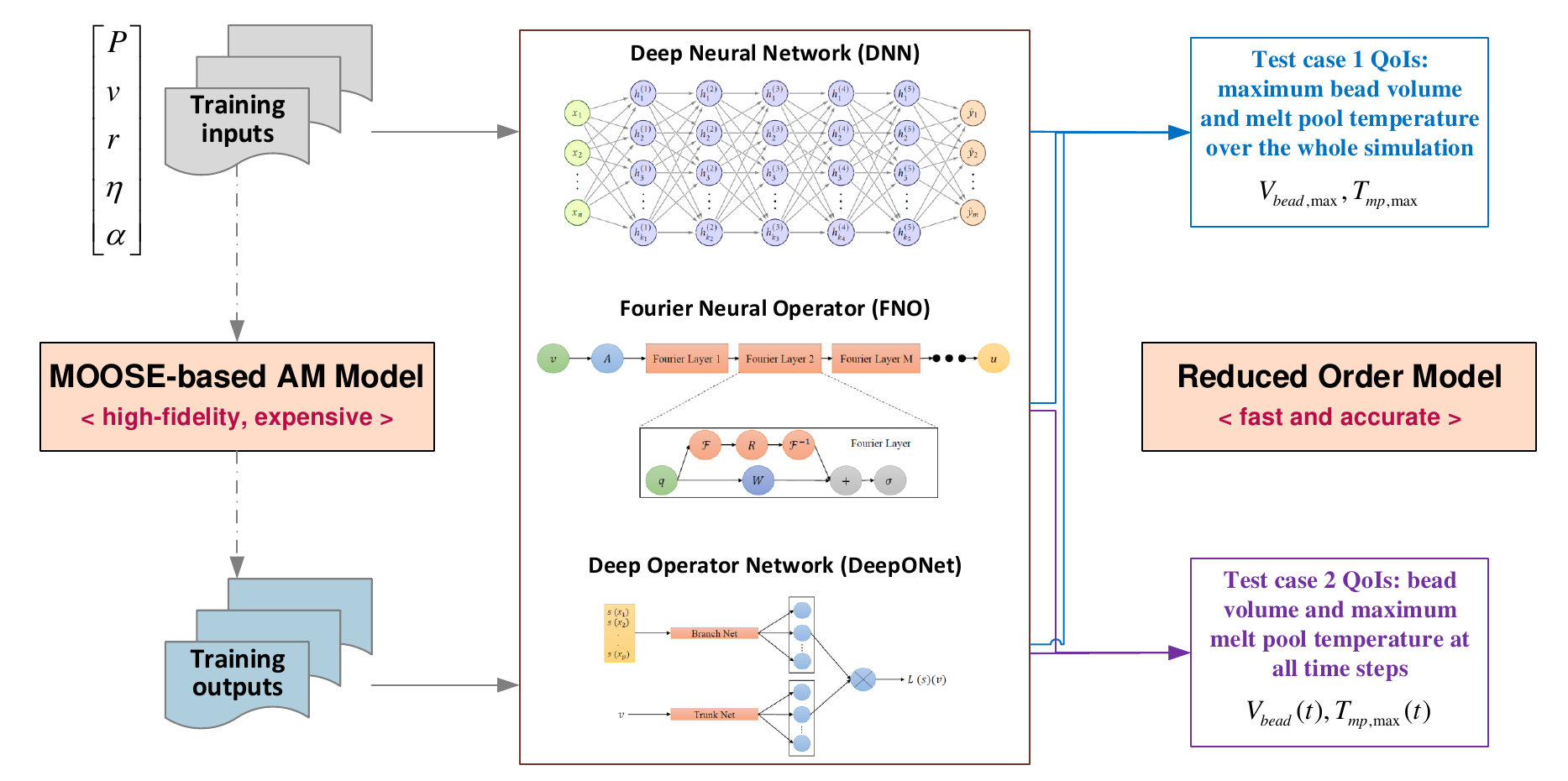}
    \caption[]{ROM workflow.}
    \label{fig:ROMs-workflow-demo}
\end{figure}

For notational convenience, $V_{\text{bead,max}}$ and $T_{\text{mp,max}}$ are used in the rest of this paper to represent the maximum values of the bead volume and melt pool temperature, respectively, over the whole simulation, whereas $V_{\text{bead}} (t)$ and $T_{\text{mp,max}} (t)$ are the time-dependent QoIs. The workflow of the ROMs employed in this study is illustrated in Figure \ref{fig:ROMs-workflow-demo}. To summarize, in the first case, we extracted $V_{\text{bead,max}}$ and $T_{\text{mp,max}}$ for each sample simulation. This dataset was used to train the three ML algorithms (i.e., DeepONet, FNO, and DNN) in order to test their abilities to predict the scalar QoIs. In the second case, only the OL-based methods (i.e., DeepONet and FNO) were used to build the ROMs for the two time-dependent responses, $V_{\text{bead}} (t)$ and $T_{\text{mp,max}} (t)$. Note that from the time series QoIs in the second case we can easily obtain the scalar QoIs by taking the maximum values over the time series. The first case is included only to demonstrate the applicability and superiority of the two OL methods over conventional DNN-based ROMs.

\section{Methodologies}
\label{section:Methodology}

In this work, both OL methods (FNO and DeepONet) use NNs as the main building components. Here, we give a brief introduction of FNO and DeepONet.

\subsection{Fourier Neural Operator}

In the field of scientific ML, the classical use of NNs is to learn mappings in finite-dimensional spaces for regression or classification tasks. When applying the standard networks to solve differential equations, the solution's spatial domain is first discretized, then the training data are generated by applying conventional numerical solvers to a single configuration (or instance) of the PDEs. If a standard NN is trained on a certain dataset, it may not generalize well to new discretizations or instances of the differential equations. This necessitates that the model be retrained with new structures and parameter optimization---a time-consuming and computationally expensive process. To resolve such issues, \cite{kovachki2021neural} proposed a new generalization of NNs, called \textit{neural operators}. Neural operators are essentially a collection of NNs with non-linear activation functions that enable approximation of any continuous operator. Neural operators learn mappings between infinite-dimensional function spaces, allowing them to directly learn the operator instead of the solution instances of a PDE. They can directly map the input function to the output function while also making both these functions discretization-invariant. The training process is executed only once, and the model can predict any input data generated from different instances of the PDEs (e.g., by changing certain coefficients, initial/boundary conditions, or the forcing term). In brief, conventional data-driven ROMs try to learn the input-output mapping for one instance of the differential equations, whereas a neural operator tries to learn an operator capable of representing many instances.

Consider the task of approximating a nonlinear operator $\mathcal{L} (h(x)) = u(x)$ for a differential equation. In this study, the input function $h(x)$ is the Gaussian heat source defined in Equation (\ref{eqn:Gaussian-point-heat-source}), whereas the solution function (or the solution $u(x)$) is the time-dependent temperature ($T$) in Equation (\ref{eqn:heat-conduction}). Denote a bounded open set $D \in \mathbb{R}^{d}$ in which the function spaces of the solution ($\mathcal{U}$) and input ($\mathcal{H}$) take the values $\mathbb{R}^{d_{u}}$ and $\mathbb{R}^{d_{h}}$. The training data are arranged into input-output pairs that represent their function spaces. Consider a dataset $\{ h_{i}, u_{i} \}_{i = 1}^{N}$, where $h_{i}$ is a sample of the input function and $u_{i} =\mathcal{L} ( h_{i})$ is the corresponding solution. This dataset includes the heat source parameters listed in Table (\ref{table:AM-process-variables}) , as well as the QoIs. It is generated by a computational model---in this work, the MOOSE-based AM model. The goal is to train a data-driven ROM of the computational model that approximates $\mathcal{L}$ with $\mathcal{L}_{\theta}$, where $\theta$ denotes the ROM's parameters (e.g., weights and bias in a NN).

The input function $h (x)$ is first lifted to a higher dimensional representation $q (x)$ by a fully connected NN, denoted as $\mathcal{A}$. $q (x)$ is updated through the layers iteratively $q_{0} \rightarrow q_{1} \rightarrow \cdots \rightarrow q_{M}$. $q_{0} (x) = \mathcal{A}\left( h (x) \right)$ is first computed, then each iteration is updated using a composition of a non-local integral operator $\mathcal{K}$ and a local, nonlinear activation function $\sigma$. $q_{M} (x)$ is then projected to the output space via local transformation. Such iterative updates are defined as follows:
\begin{equation}    \label{eqn:OL-iterative-updates}
    q_{m+1} (x) = \sigma \left( W q_{m} (x) +   \left( \mathcal{K} (h; \theta) q_{m} \right)  (x) \right), \: \forall x \in D
\end{equation}
where $W$ is a linear transformation, $\sigma$ is a non-linear activation function evaluated in a point-wise manner, and $\mathcal{K}$ is an integral kernel operator defined as follows:
\begin{equation}    \label{eqn:integral-kernel-operator}
     \left( \mathcal{K} ( h; \theta) q_{m} \right)  (x) = \int_{D} \kappa \left( x, y, q(x) ; \theta \right) q (y) \: dy, \: \forall x \in D
\end{equation}
where $\kappa$ is a NN parameterized by $\theta \in \Theta_{\mathcal{K}}$ and learned from data.

Despite the integral operator being linear, it can learn highly complex non-linear operators by using linear integral operators with non-linear activation functions. Neural operator layers can take the form of a graph neural operator, lower-rank neural operator, and FNO, as detailed in \cite{kovachki2021neural}. The present work uses FNO, based on which the kernel in Equation \ref{eqn:integral-kernel-operator} is defined in the Fourier space \cite{li2020fourier}. By requiring that $\kappa (x,y) = \kappa (x - y)$, the integral operator becomes a convolution operator (defined in the Fourier space) that can be computed using the fast Fourier transform technique (see \cite{li2020fourier} for the full derivation). Let $\kappa$ be periodic, the Fourier series expansion will produce discrete frequency modes that can be truncated at a certain number to obtain the desired level of accuracy at an acceptable computational cost. Both input and output functions must be defined in the same domain and employ the same discretization scheme, as the evaluations are carried out in point-wise fashion for an equally spaced mesh. Thus, FNO will learn the mapping from the discretization of $h$ to the discretization of $u$ in the same equally spaced mesh.

\begin{figure}[!htb]
	\centering
	\captionsetup{justification=centering}
	\includegraphics[width=0.7\textwidth]
        {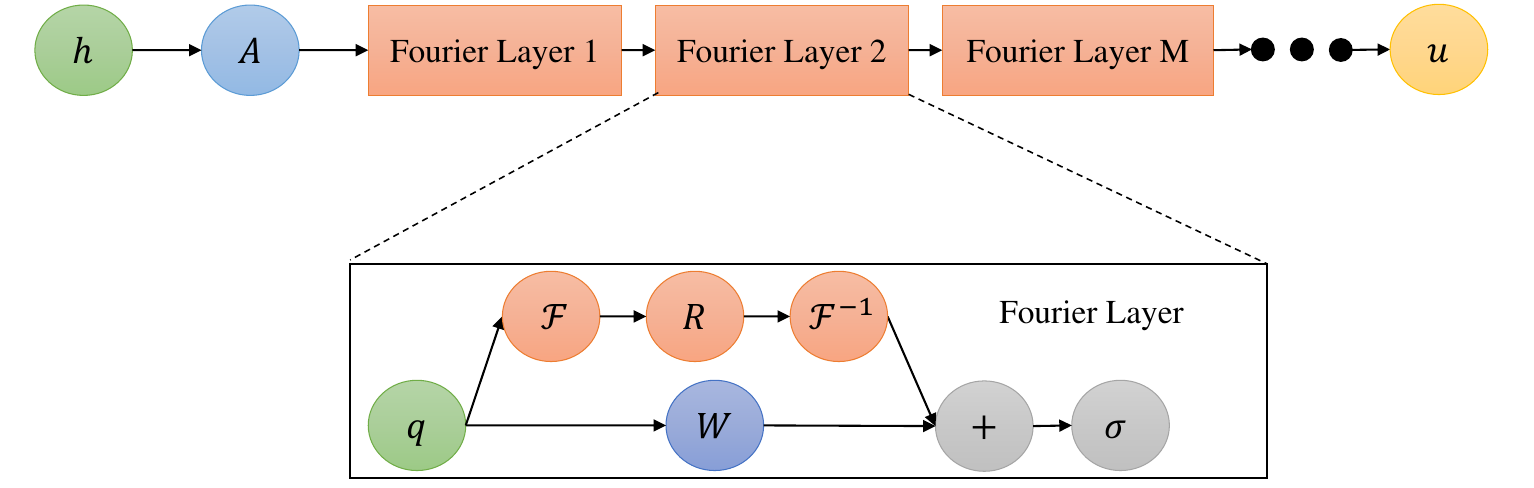}
	\caption[]{Illustration of the FNO structure. Figure adapted from \cite{li2020fourier}.}
	\label{fig:neural-operator-FNO-demo}
\end{figure}

In each layer shown in Figure \ref{fig:neural-operator-FNO-demo}, a Fourier transform is implemented for the transformed input $q$. Next, the higher-order modes are filtered out before an inverse Fourier transform is performed. A NN projects the output of the last Fourier layer back to the target dimension. To make a prediction, the trained FNO model computes the Fourier coefficients of the desired QoIs. Those coefficients are converted back into the original data domain via the inverse Fourier transform. For the present work, to simulate time-dependent QoIs, we modified the structure of the Fourier layers to accept input parameters that vary with time. Since the input parameter values in our case study remain constant over time, they are repeated over 200 time steps for each sample. Each input vector is represented by a specified number of modes. This modification enables simulation of time-dependent input parameters by merely replacing constant values with new time series for each parameter in a given sample. Example FNO modifications are given in \cite{lu2022comprehensive}.


\subsection{Deep Operator Network}

\begin{figure}[!htb]
	\centering
	\captionsetup{justification=centering}
	\includegraphics[width=0.7\textwidth] 
        {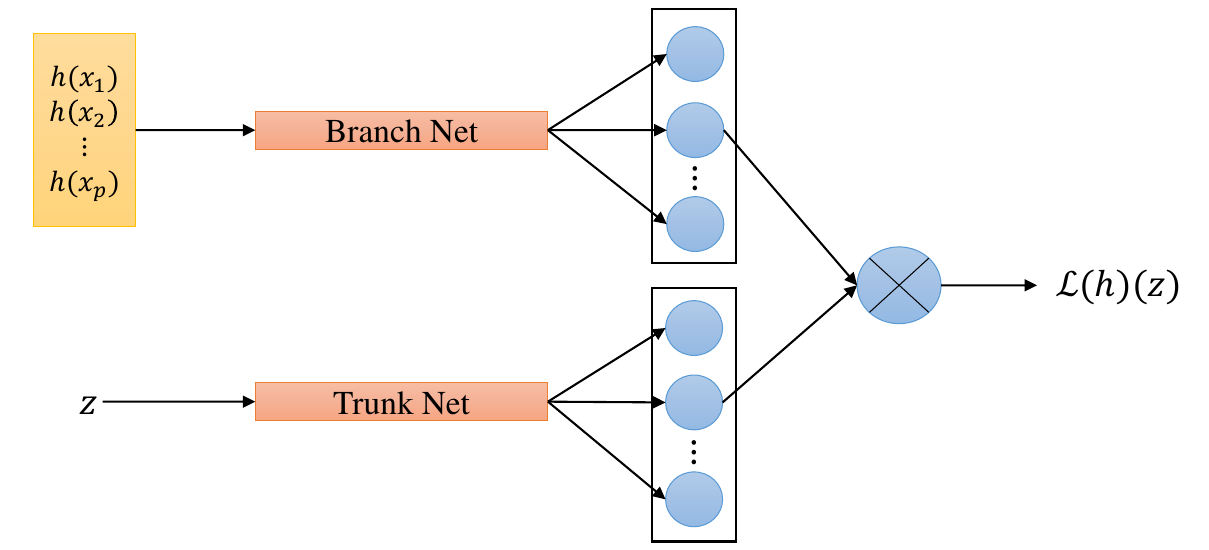}
	\caption[]{DeepONet structure. Figure adapted from \cite{lu2021learning}.}
	\label{fig:neural-operator-DeepONet-demo}
\end{figure}

DeepONet was originally proposed in \cite{lu2021learning}, based on the fact that NNs can approximate any continuous nonlinear operator, as per the universal approximation theorem. The authors expanded this concept to DNNs because they have more expressivity than single-layer NNs. We will use the same notation for the operator $\mathcal{L}$ that maps an input function $h$ to an output function $\mathcal{L}(h)$. The values of $\mathcal{L}(h)$ are computed at all points in the spatial and/or temporal domain $z$, such that $\mathcal{L}(h)(z)$ is a real number. Therefore, the DeepONet takes two inputs, $h$ and $z$, and the outputs are denoted as $\mathcal{L}(h)(z)$. The input function $h$ at $p$ locations (also referred to as sensors) is encoded into a DNN called the branch net, and the sensor vector $z$ is encoded into another DNN called the trunk net. The training data consist of the input function vector $[h (x_{1}), h (x_{2}), \cdots, h (x_{p})]$ evaluated at a fixed number of sensors $\{x_1, \ldots,x_p\}$, while the output $\mathcal{L}(h)(z)$ is evaluated at certain sensor locations not necessarily identical to those used for the input function. In this study, the input function is the Gaussian heat source, and the outputs are the solution of the thermal model plus the bead volume of the printed material. The DeepONet has two basic architectures \cite{lu2021learning}: stacked and unstacked. In the present work, we used the unstacked architecture (see Figure \ref{fig:neural-operator-DeepONet-demo}), which is composed of one branch net and one trunk net. (The stacked one consists of multiple branch nets and one trunk net.) We employed the unstacked DeepONet because it is less expensive to train. Moreover, as indicated in \cite{lu2021learning}, its generalization capability is better, since it produces smaller test errors than the stacked architecture.

DeepONet employs the same concept as other OL algorithms---namely, learning linear and nonlinear operators by using simulation data collected from a numerical solver of the differential equation. It can achieve a lower generalization error in comparison to fully connected NNs, since it uses two sub-networks combined by a dot product to generate the output. In the DeepONet model, the branch net takes the source term in the governing equation as an input function, and the trunk net takes the vector of sensors as input. In our MOOSE model, the source term is the Gaussian heat source, which depends on five parameters. Thus, the five input parameters are given to the branch net, and the vector of time steps is given to the trunk net. This modification to the DeepONet structure allowed us to build ROMs for time-dependent QoIs.


\section{Data generation and preprocessing}
\label{section:Data-generation-and-pre-processing}

Based on the nominal values and lower/upper bounds of the five input parameters---$P$, $\alpha$, $\eta$, $r$, and $v$ in Table \ref{table:AM-process-variables}---we used Latin hypercube sampling to generate 500 uniformly distributed samples of the parameters, within the specified intervals. The MOOSE-based AM model (see Section \ref{section:MOOSE-AM-model}) was executed for each sample in order to produce the QoIs. The output data contain the time-dependent bead volume and maximum melt pool temperature for all 200 time steps. Note that given the fixed line scan length, the line scan times in the simulation will differ based on the scanning speed $v$. Therefore, for comparing the results, we used different time step sizes for different scanning speeds in order to reach the same total number of 200 time steps for every sample. This is enough time steps to ensure that the sample with the smallest scanning speed $v$ has an acceptable time step size. 

For preprocessing the training data, all samples that produced non-melting conditions were removed from the training stage, as these simulations were considered outliers in the training data. Each sample in the training data should correspond to an actual DED process in which the powder material is melted and shaped into the desired geometry. If the non-melting samples are retained in the training data, the trained ROMs exhibit considerably larger errors. A total of 18 samples was removed from the original dataset. Since the volume and temperature scales were different, the output matrix was standardized using scikit-learn library's \cite{scikit-learn} Standard Scaler function, thereby improving the ML algorithm accuracy and making the learning process easier for each test case. For training, validation, and testing, the dataset was divided into 80$\%$, 10$\%$, and 10$\%$, respectively.

\section{Results and Discussions}
\label{section:Results}

The results of the ROM simulations for test case 1 are presented in Section \ref{section:results-maximum-QoIs}. Those for the second test case are presented in Section \ref{section:results-time-series-QoIs}.

\subsection{Results for the maximum QoIs test case}
\label{section:results-maximum-QoIs}

This section is divided into three subsections. In the first, we present the results of a systematic hyperparameter tuning process, and show the results from six groups of models in order to test the impact of certain hyperparameters on prediction accuracy. In the second subsection, we present all the ROM predictions for the maximum QoIs test case and make a comprehensive comparison in terms of accuracy, error distribution, and computational efficiency. In the last subsection, we present the results of a global sensitivity analysis (SA) that was conducted  using the fast-running ROMs to investigate the sensitivity of the two scalar QoIs with respect to the five input parameters.

\subsubsection{ROM hyperparameter sensitivity analysis}
\label{section:results-hyperparameter}

Hyperparameter tuning is generally needed---using the validation dataset to find the ``best'' hyperparameters for the ML models. We implemented the grid search method to select the hyperparameter values that maximize prediction accuracy. In this section, we present some of the hyperparameter tuning results to show how several different hyperparameters impact the overall accuracy of different ROMs. More specifically, we examined the effects, or sensitivity, of adjusting the following hyperparameters: (1) the number of neurons and hidden layers for DNN, (2) the number of neurons and hidden layers within both the branch and trunk networks for DeepONet, and (3) the number of modes and Fourier layers for FNO.

Table \ref{table:test-models-specs} shows the definitions of the six sets/groups of models to be compared. For model groups 1--3, the number of neurons was changed for both DeepONet and DNN, and the number of modes for FNO were varied while keeping the number of layers and all other hyperparameters fixed. For model groups 4--6, the number of layers for all the ROMs was varied while keeping both the number of neurons in each layer in DeepONet/DNN and the number of FNO modes fixed. For DeepONet, the values $a/b$ denote the $a$ and $b$ hidden layers used in the branch net and trunk net, respectively.

\begin{table}[htbp!]
\footnotesize
  \centering
  \caption{Hyperparameters used in the six sets/groups of ROMs. In the DeepONet column, the first number in each row represents the number of neurons per layer, while the second pair of numbers corresponds to the number of layers in both the branch and trunk networks. In the FNO column, the top and bottom values are the number of modes and the number of Fourier layers, respectively. In the DNN column, the top row represents the number of neurons in each hidden layer, while the number on the bottom is the number of hidden layers.}
  \label{table:test-models-specs}
  \begin{tabular}{c c c c c}  
  \toprule
  Model sets/groups  &  Hyperparameters & DeepONet & FNO & DNN \\
  \midrule
  \multirow{2}{*}{1} & Neurons/Modes & \multirow{1}{*}{100} & \multirow{1}{*}{1} & \multirow{1}{*}{100, 150, 200, 150, 100} \\
                     & Layers & \multirow{1}{*}{2/1} & \multirow{1}{*}{4} & \multirow{1}{*}{5} \\
  \midrule
  \multirow{2}{*}{2} & Neurons/Modes  & \multirow{1}{*}{200} & \multirow{1}{*}{5} & \multirow{1}{*}{150, 200, 250, 200, 150} \\
                     & Layers & \multirow{1}{*}{2/1} & \multirow{1}{*}{4} & \multirow{1}{*}{5} \\

  \midrule
  \multirow{2}{*}{3} & Neurons/Modes  & \multirow{1}{*}{300} & \multirow{1}{*}{9} & \multirow{1}{*}{250, 300, 350, 300, 250} \\
                     & Layers & \multirow{1}{*}{2/1} & \multirow{1}{*}{4} & \multirow{1}{*}{5} \\

  \midrule
  \multirow{2}{*}{4} & Neurons/Modes  & \multirow{1}{*}{150} & \multirow{1}{*}{1} & \multirow{1}{*}{300, 300} \\
                     & Layers & \multirow{1}{*}{3/2} & \multirow{1}{*}{1} & \multirow{1}{*}{2} \\

  \midrule
  \multirow{2}{*}{5} & Neurons/Modes & \multirow{1}{*}{150} & \multirow{1}{*}{1} & \multirow{1}{*}{300, 300, 300} \\
                     & Layers & \multirow{1}{*}{4/3} & \multirow{1}{*}{2} & \multirow{1}{*}{3} \\

  \midrule
  \multirow{2}{*}{6} & Neurons/Modes & \multirow{1}{*}{150} & \multirow{1}{*}{1} & \multirow{1}{*}{300, 300, 300, 300} \\
                     & Layers & \multirow{1}{*}{5/4} & \multirow{1}{*}{3} & \multirow{1}{*}{4} \\
                     
  \bottomrule
  \end{tabular}
\end{table}

\begin{figure}[!htb]
	\centering
	\captionsetup{justification=centering}
	\includegraphics[width=0.8\textwidth] 
        {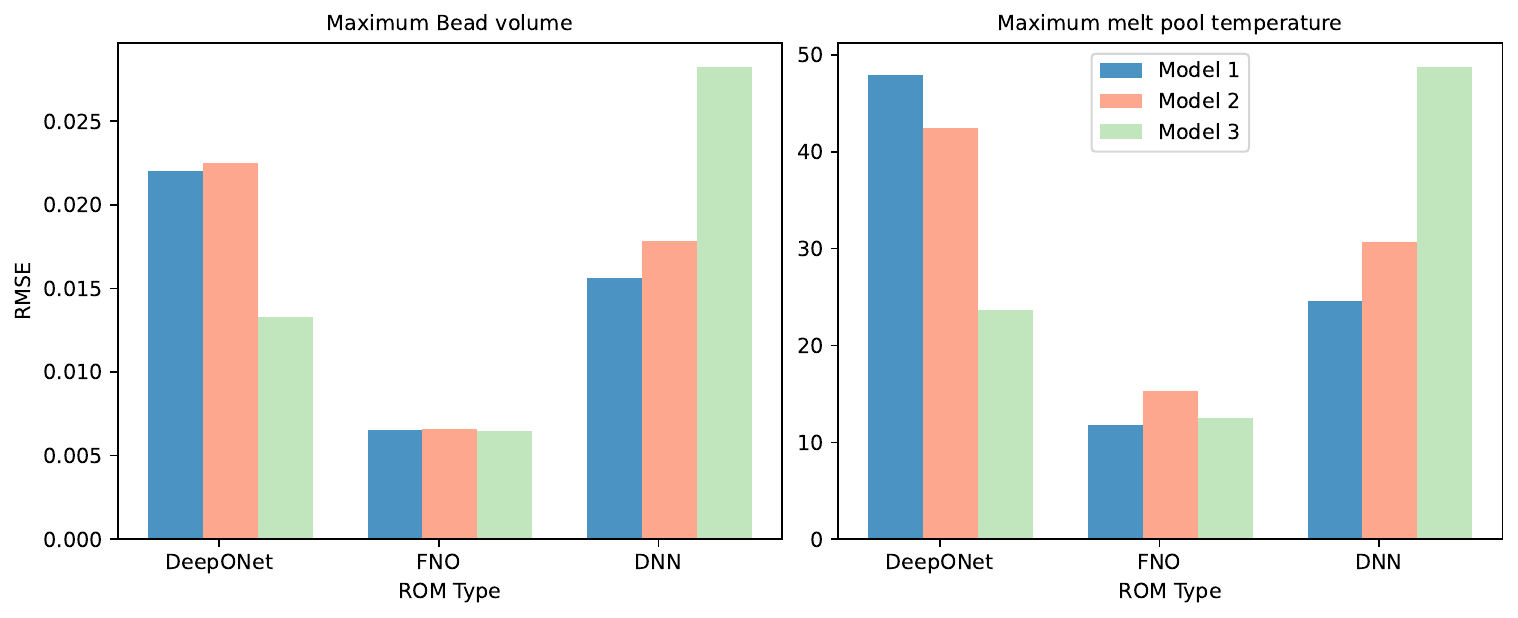}
	\caption[]{Effects of the number of neurons (DeepONet/DNN) and modes (FNO) on ROM accuracy, based on comparing the RMSEs from model groups 1--3.}
	\label{fig:hp-sensitivity-neurons}
\end{figure}

\begin{figure}[!htb]
	\centering
	\captionsetup{justification=centering}
	\includegraphics[width=0.8\textwidth] 
        {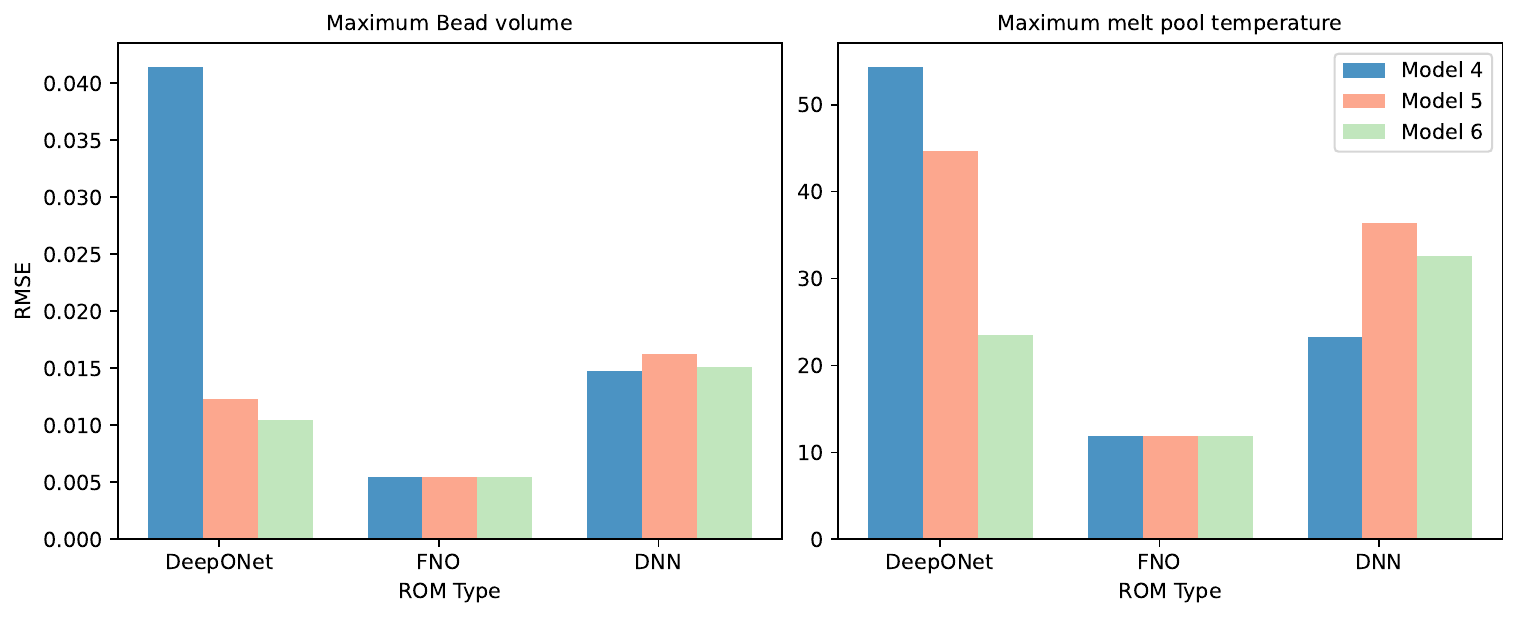}
	\caption[]{Effects of the number of layers (DeepONet/DNN) and Fourier layers (FNO) on ROM accuracy, based on comparing the RMSEs from model groups 4--6.}
	\label{fig:hp-sensitivity-layers}
\end{figure}

\begin{figure}[!htb]
	\centering
	\captionsetup{justification=centering}
	\includegraphics[width=0.8\textwidth] 
        {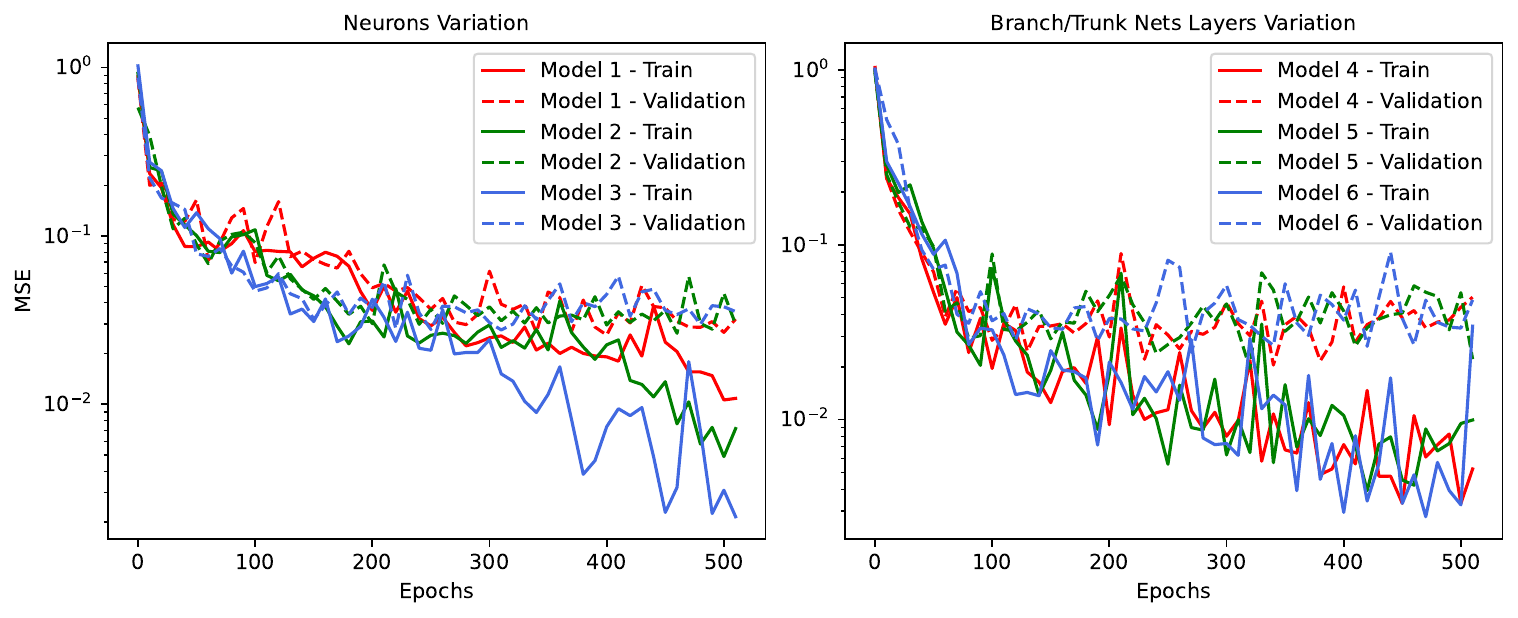}
	\caption[]{Training and validation losses of the DeepONet-based ROMs.}
	\label{fig:DeepONet-learning-curve}
\end{figure}

\begin{figure}[!htb]
	\centering
	\captionsetup{justification=centering}
	\includegraphics[width=0.8\textwidth] 
        {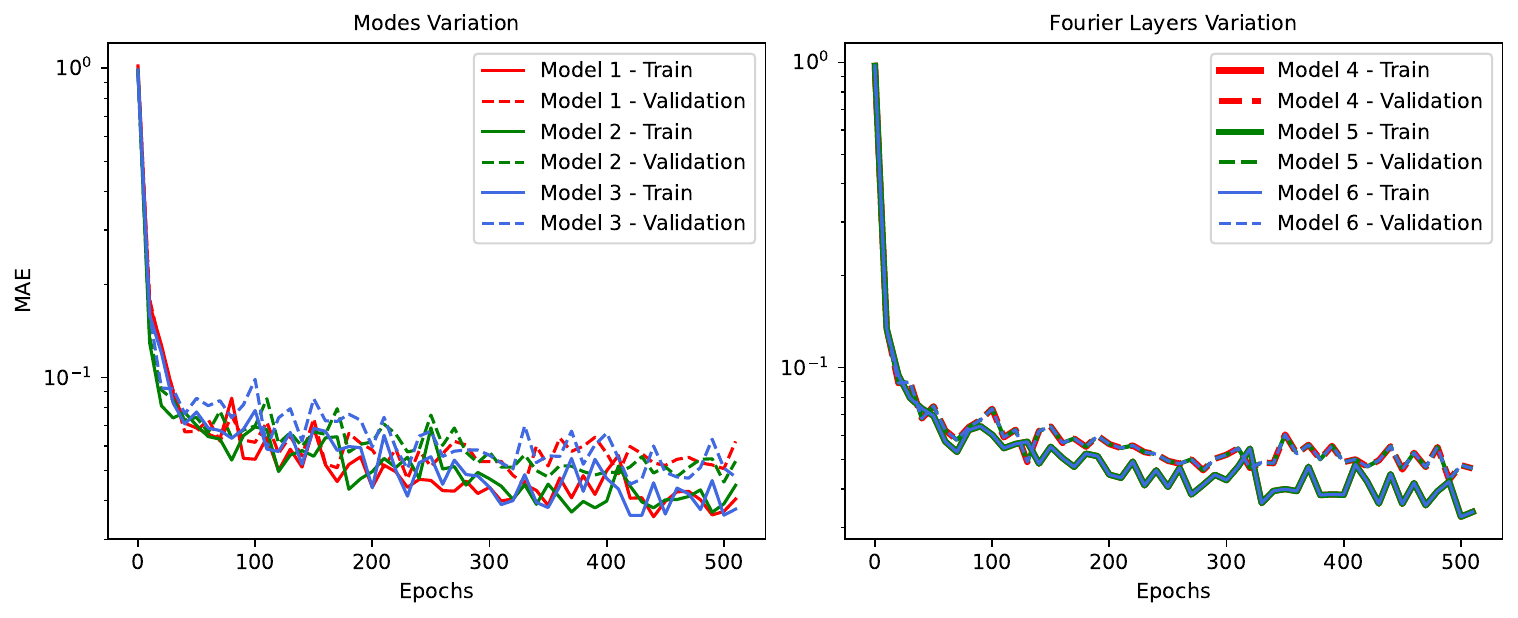}
	\caption[]{Training and validation losses of the FNO-based ROMs.}
	\label{fig:FNO-learning-curve}
\end{figure}

\begin{figure}[!htb]
	\centering
	\captionsetup{justification=centering}
	\includegraphics[width=0.8\textwidth] 
        {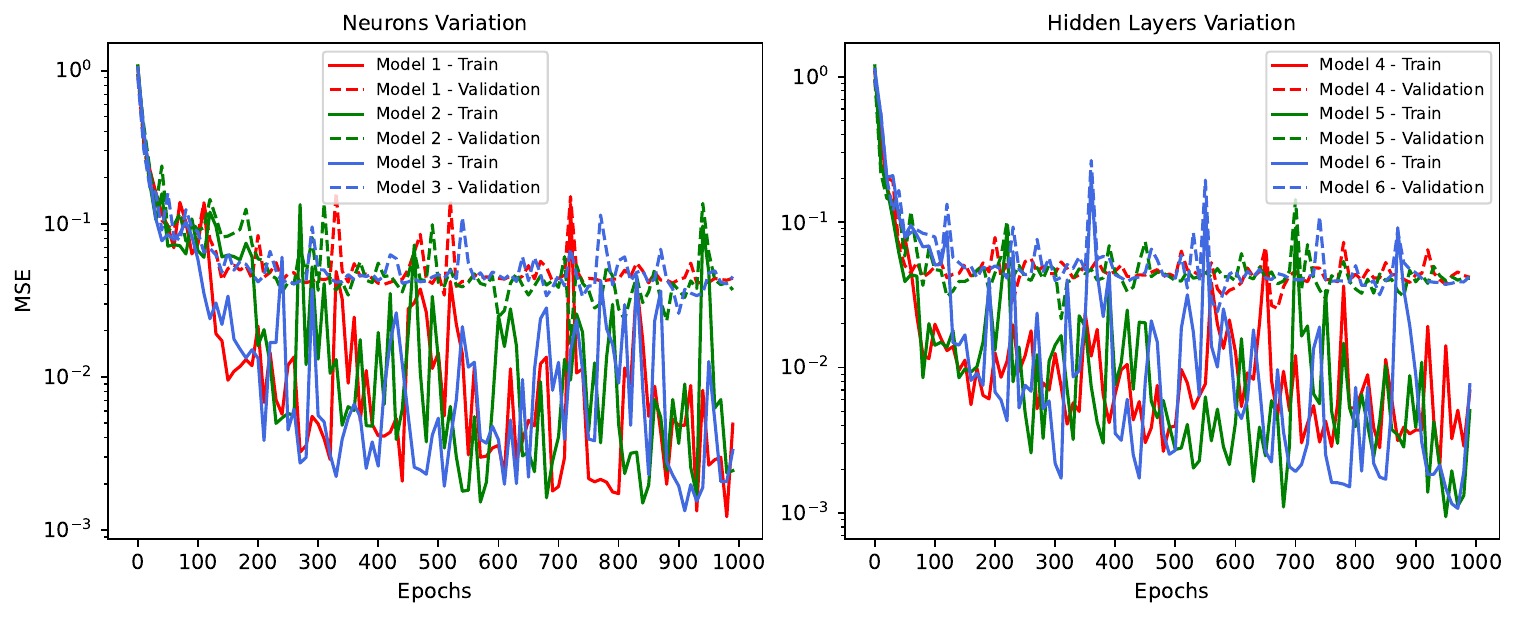}
	\caption[]{Training and validation losses of the DNN-based ROMs.}
	\label{fig:DNN-learning-curve}
\end{figure}

The accuracy of the ROMs was evaluated based on their prediction RMSEs over the test dataset. Figure \ref{fig:hp-sensitivity-neurons} shows the RMSEs by varying the number of neurons (DeepONet/DNN) and modes (FNO). Figure \ref{fig:hp-sensitivity-layers} shows the RMSEs by changing the number of layers (DeepONet/DNN) and Fourier layers (FNO). Finally, Figures \ref{fig:DNN-learning-curve} and  \ref{fig:DeepONet-learning-curve} present the DeepONet/DNN training and validation losses in terms of mean squared errors, while Figure \ref{fig:FNO-learning-curve} presents the FNO training and validation losses in terms of mean absolute error.

The DeepONet-based ROMs learn better with increasing neurons as the prediction RMSE and training mean squared error are reduced, as shown in the left-hand plots of Figures \ref{fig:hp-sensitivity-neurons} and \ref{fig:DeepONet-learning-curve}. However, the validation loss is insensitive to the number of neurons and hidden layers (see Figure \ref{fig:DeepONet-learning-curve}). Note that the $V_{\text{bead,max}}$ prediction RMSE slightly increased for model 2 and then decreased for model 3, due to the higher variance in the DeepONet prediction error for $V_{\text{bead,max}}$, as will be shown later. Increasing the number of hidden layers for both the branch and trunk nets enhances the performance for both QoIs, as shown in Figure \ref{fig:hp-sensitivity-layers}. The $V_{\text{bead,max}}$ is shown to be more sensitive to the number of layers than to the neurons, as the reduction in RMSE is larger in the latter case. For $T_{\text{mp,max}}$, DeepONet is equally sensitive for both hyperparameters.

For FNO-based ROMs, neither $V_{\text{bead,max}}$ nor $T_{\text{mp,max}}$ are sensitive to the modes of the Fourier series and the number of Fourier layers, as shown in Figures \ref{fig:hp-sensitivity-neurons} and \ref{fig:hp-sensitivity-layers}. Only small variations can be observed for $T_{\text{mp,max}}$ when changing the number of modes. This is primarily because the FNO predictions for certain samples may turn oscillatory with a reduced number of modes---a phenomenon that does not align with physical correctness. Per the training and validation losses given in Figure \ref{fig:FNO-learning-curve}, neither the number of modes nor the number of Fourier layers affect the learning process for both QoIs. The fact that the FNO training and validation losses are much smaller than the other two methods indicates negligible over-fitting, giving it the lowest generalization error among all the ROMs tested for the test dataset.

For the conventional DNN-based ROMs, the predicted RMSE increases as the number of neurons increases, as the network becomes more complex and starts to memorize specific details of the training data rather than simply generalizing for unseen datasets (see Figure \ref{fig:hp-sensitivity-neurons}). Figure \ref{fig:hp-sensitivity-layers} shows that when more hidden layers are added, the DNN prediction RMSE first increases and then decreases for both QoIs, for the same reason mentioned earlier in regard to the DeepONet model's sensitivity to neurons. The training and validation losses shown in Figure \ref{fig:DNN-learning-curve} are not reduced with increasing model complexity. Overall, the DNN model should be simple in order to achieve good prediction accuracy. This is mainly because of the limited amount of training data available.

\subsubsection{Comparison of ROM results}

\begin{figure}[!htb]
	\centering
	\captionsetup{justification=centering}
	\includegraphics[width=0.9\textwidth] 
        {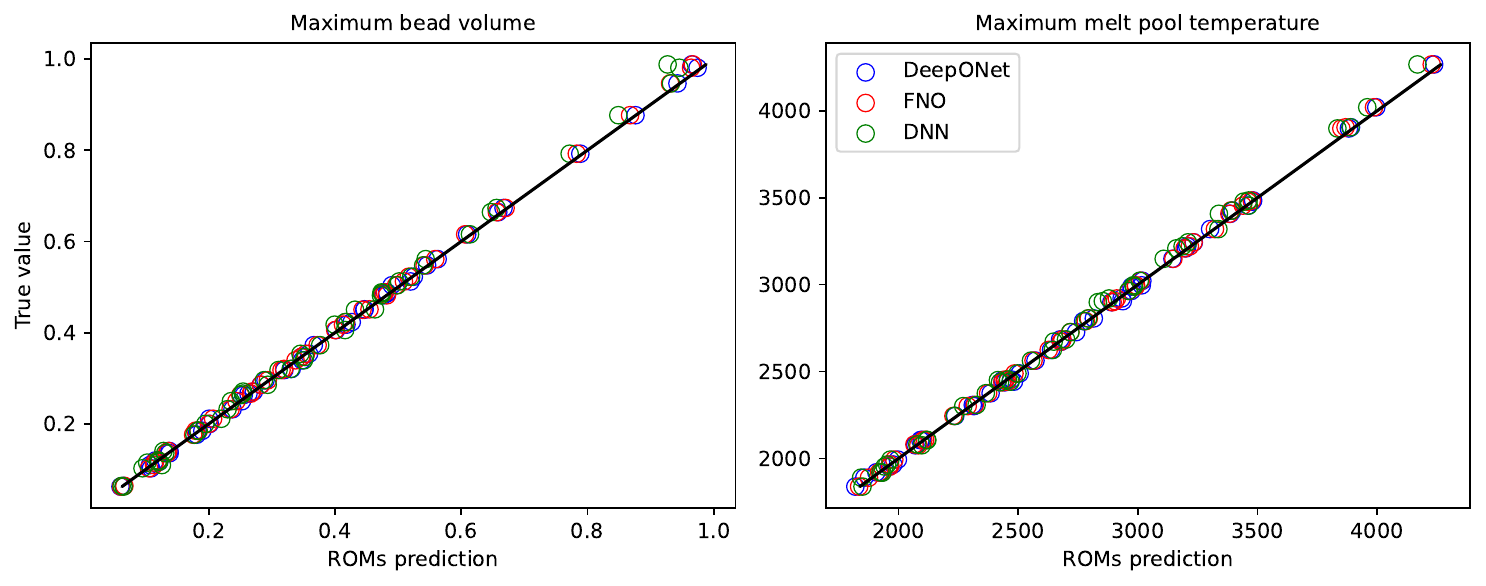}
	\caption[]{Comparison of ROM predictions for $V_{\text{bead,max}}$ and $T_{\text{mp,max}}$.}
	\label{fig:ROMs-pred-max-vol-temp}
\end{figure}

Figure \ref{fig:ROMs-pred-max-vol-temp} presents the three ROMs' predictions for $V_{\text{bead,max}}$ and $T_{\text{mp,max}}$, as compared to the true values found in the test dataset. The OL algorithms' predictions for all test samples agree very well with the actual values of those test samples. The DNN predictions show good agreement as well, though a few test samples reflect slightly larger deviations. The DeepONet model training time was 395.8 seconds, and the FNO model took 219.9 seconds to finish the optimization process. In contrast, the DNN model had a faster training time: 78.7 seconds. From a computational efficiency point of view, it is the most efficient ROM. However, its predictions are less accurate compared to the two OL algorithms. For a better quantitative comparison, Table \ref{table:FNO-DNN-performance_comparison} presents the RMSE, as well as the coefficient of determination ($R^2$), which represents how well each ROM approximates the test samples. The RMSE values for DeepONet and FNO are close to each other; compared to DNN, they are almost $50\%$ and $250\%$ less for both $V_{\text{bead,max}}$ and $T_{\text{mp,max}}$, respectively. The $R^2$ values for the FNO and DeepONet models are closer to 1.0 than are those of the DNN, indicating that the OL approaches have a better fit to the true model than does the DNN approach. The performances of both OL algorithms are comparable, with FNO showing a slight edge as a result of being faster and having smaller RMSE values.

\begin{table}[htbp!]
        \footnotesize
        \centering
        \caption{Comparison of the performance of the three ROMs.}
        \label{table:FNO-DNN-performance_comparison}
        \begin{tabular}{c c c c c c c}        
        \toprule        
        \multirow{1}{*}{Scalar QoIs}  & \multicolumn{2}{c}{FNO} & \multicolumn{2}{c}{DeepONet} & \multicolumn{2}{c}{DNN} \\        
        \cmidrule{2-7}  &  RMSE   & $R^{2}$  & RMSE & $R^{2}$ & RMSE & $R^{2}$ \\        
        \midrule
        $V_{\text{bead,max}}$  & $6.61 \times 10^{-3}$ & 0.9993    &  $6.07 \times 10^{-3}$ & 0.9993 & $1.51 \times 10^{-2}$ & 0.9961 \\        
        \midrule
        $T_{\text{mp,max}}$ & 15.3 &  0.9994 & 15.9 &  0.9993 & 32.6 & 0.9972  \\        
        \bottomrule
        \end{tabular}
\end{table}

\begin{figure}[!htb]
	\centering
	\captionsetup{justification=centering}
	\includegraphics[width=0.9\textwidth] 
        {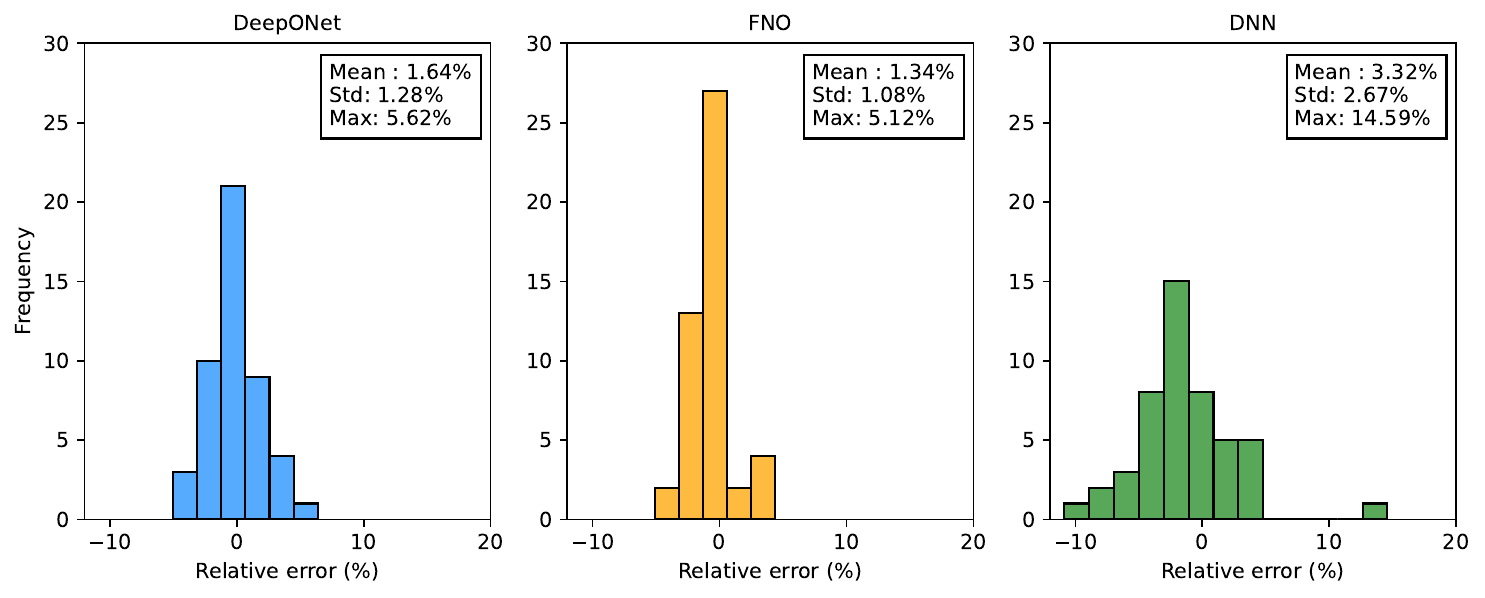}
	\caption[]{Relative error distribution of $V_{\text{bead,max}}$ predictions.}
	\label{fig:ROMs-RE-hist_bead-volume}
\end{figure}

\begin{figure}[!htb]
	\centering
	\captionsetup{justification=centering}
	\includegraphics[width=0.9\textwidth] 
        {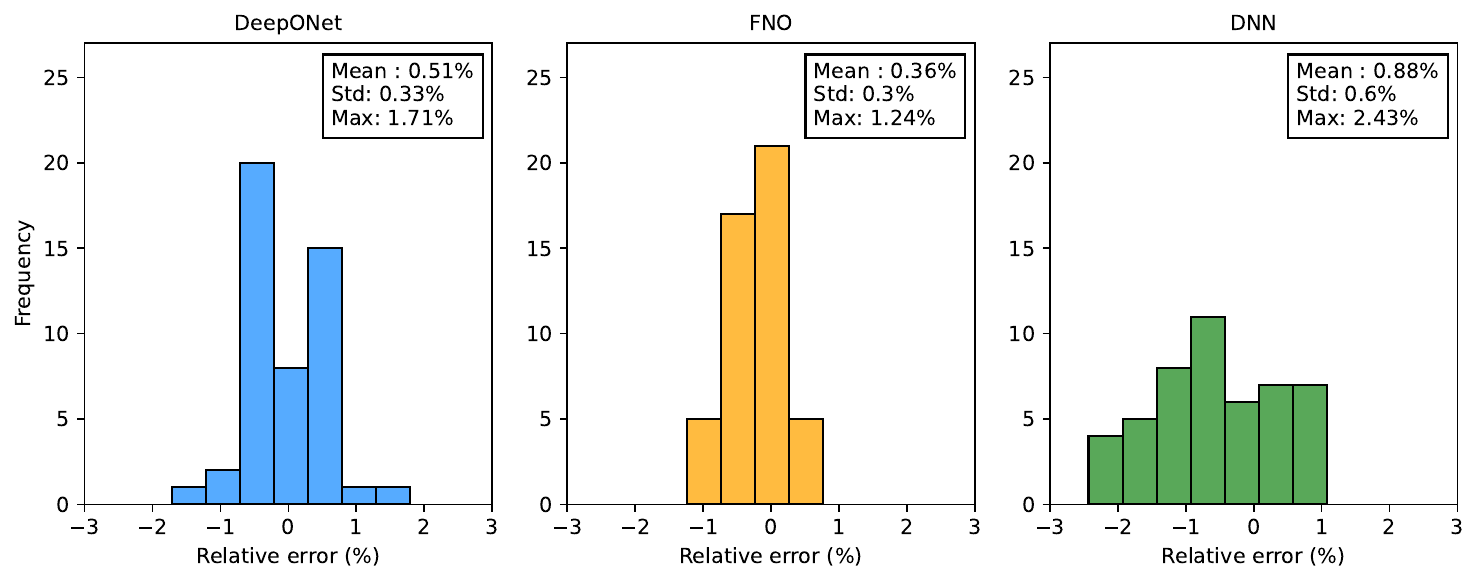}
	\caption[]{Relative error distribution of $T_{\text{mp,max}}$ predictions.}
	\label{fig:ROMs-RE-hist-max-temp}
\end{figure}

\begin{figure}[!htb]
	\centering
	\captionsetup{justification=centering}
	\includegraphics[width=0.9\textwidth] 
        {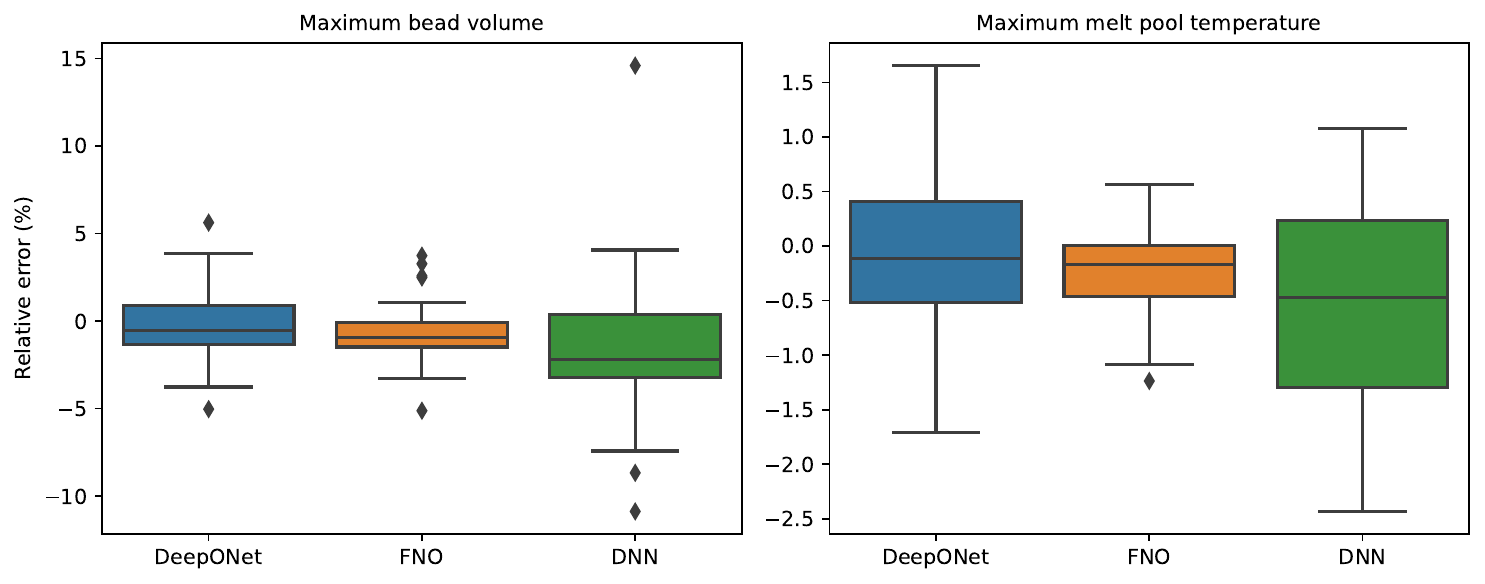}
	\caption[]{Relative error box plot.}
	\label{fig:ROMs-RE-max-box-plot}
\end{figure}

Figures \ref{fig:ROMs-RE-hist_bead-volume} and \ref{fig:ROMs-RE-hist-max-temp} demonstrate, for all test samples, the distribution of the relative error percentage for both QoIs. The histogram of the DeepONet model error for $V_{\text{bead,max}}$ shows that the bar of higher frequency lies around small error values. This is also observed for FNO. The DNN error is larger and has a wider spread (higher variance). In the temperature response histograms, DeepONet and FNO have smaller variances than $V_{\text{bead,max}}$ and thus, a larger possibility of predicting outliers that will increase the RMSE value for volume predictions over temperature, as observed earlier in section \ref{section:results-hyperparameter}. To afford a better understanding of the error distribution, Figure \ref{fig:ROMs-RE-max-box-plot} shows a box plot of the test sample error. It displays a five-number summary of a dataset; namely, the minimum, first quartile, median, third quartile, and maximum values. For $V_{\text{bead,max}}$, the DeepONet and FNO boxes are smaller than the DNN box, since most of the error values are concentrated around the mean. The FNO box has outliers points that correspond to larger errors and lay in the leftmost side of the histogram. All the ROMs show the maximum error value as an outlier, since it is larger than the third quantile plus 1.5 in the interquartile range. As for temperature response, all the ROM test sample errors are more concentrated around the mean value. Thus, no outliers are shown for DeepONet and DNN, and only one is shown for FNO. For both QoIs, the error distribution is noticeably skewed and asymmetric, as the median line is not in the middle of the box.

\subsubsection{Global sensitivity analysis}

\begin{figure}[!htb]
	\centering
	\captionsetup{justification=centering}
	\includegraphics[width=1.0\textwidth] 
        {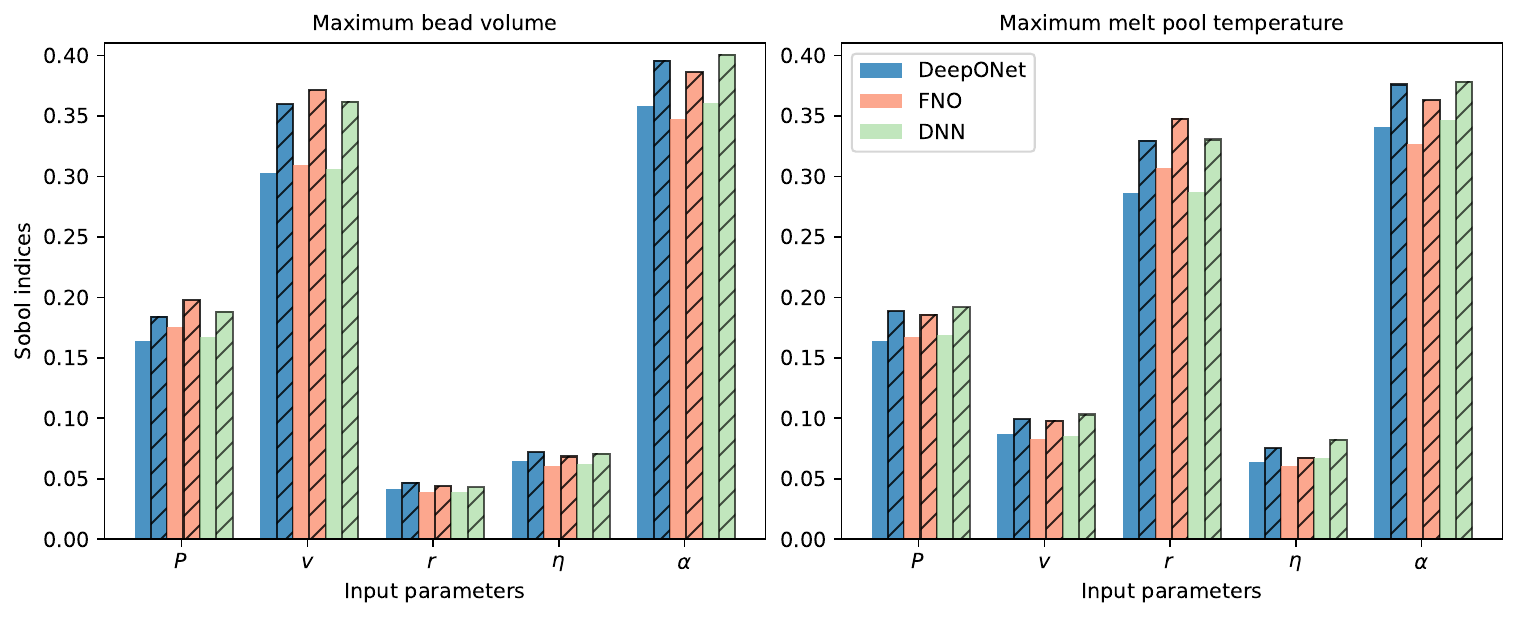}
	\caption[]{First (solid bars) and total (hatched bars) Sobol indices for both QoIs, as computed via the three ROMs.}
	\label{fig:ROMs-Sobol-S1-ST}
\end{figure}

An important study for physics-based modeling and simulation is SA, which is defined as the study of how uncertainties in the QoIs can be apportioned to various input parameters. SA ranks the input parameters according to their significance to the QoIs. Such information can be very useful in physical model development. For example, one may focus on reducing the uncertainties in the significant inputs in order to reduce the prediction uncertainties in the QoIs. One of the most popular global SA methods is to use variance-based SA to compute the Sobol indices \cite{sobol2001global} \cite{saltelli2010variance}. One primary challenge is that the computational cost is usually very high if the physical model is expensive to run. However, with fast and accurate ROMs, we can easily compute the Sobol indices in a short time. We used the Sensitivity Analysis Library in Python \cite{Iwanaga2022} \cite{Herman2017} to perform this analysis. We computed the first-order and total Sobol indices for both QoIs (see Figure \ref{fig:ROMs-Sobol-S1-ST}). All the ROMs displayed similar analysis results, and the total Sobol indices shown in the hatched bars are slightly larger than the first-order indices. We used 1000 samples of the input parameters; all the ROMs generated their predictions within a mere moment. Had we used the MOOSE physical model, it would have taken 17.3 days to run all 1000 samples. Thus, using ROMs significantly decreased the runtime. 

The first-order indices indicate the contribution of the specific input parameter to the QoIs. It thus measures the effect of varying that parameter alone. Based on Figure \ref{fig:ROMs-Sobol-S1-ST}, both QoIs are highly sensitive to the scaling factor, because this parameter has a wider interval of change and is a multiplication of the laser power. The $V_{\text{bead,max}}$ is more sensitive to the scanning speed, as a faster-moving laser will print more material and thus its dimensions will increase. On the other hand, $T_{\text{mp,max}}$ is more affected by the radius of the laser, since a smaller radius implies that more energy will be deposited in a given spot, and because higher temperature values are obtained. Both QoIs have a similar sensitivity to efficiency and laser power, since the efficiency factor will control the quality of the laser used and the power will control the energy deposited in the powder; therefore, they contribute equally to the melt pool temperature and the amount of material melted.  

The total Sobol indices of an input parameter measure the variance contributions of its interaction with any other parameters, in any order. Table \ref{table:total-Sobol-sum} lists the summation of the total Sobol over all parameters for each ROM. The fact that the summation is only slightly larger than 1.0 indicates that the interactions among all parameters are insignificant. Thus, in the SA in this work, it is unnecessary to consider the high-order interactions of the input parameters.

\begin{table}[htbp!]
        \footnotesize
        \centering
        \caption{Summation of the total Sobol indices to evaluate the effects of parameter interactions.}
        \label{table:total-Sobol-sum}
        \begin{tabular}{c c c}        
        \toprule
         ROMs & $V_{\text{bead,max}}$ & $T_{\text{mp,max}}$ \\
        \midrule
        DeepONet & 1.0577	& 1.0679 \\
        FNO   & 	1.0670 &	1.0611 \\
        DNN & 1.0638	& 1.0860 \\
        \bottomrule
        \end{tabular}
\end{table}

\subsection{Results for the time-dependent QoIs test case}
\label{section:results-time-series-QoIs}

In Section \ref{section:results-maximum-QoIs}, we demonstrated that, in building ROMs, the two OL approaches (DeepONet and FNO) offer better accuracy than conventional DNNs. In the first test case, the scalar QoIs, $V_{\text{bead,max}}$ and $T_{\text{mp,max}}$, were considered. In this section, we now present the results for the second test case, in which time-dependent responses are used as the two QoIs---namely, the bead volume $V_{\text{bead}} (t)$ and maximum melt pool temperature $T_{\text{mp,max}} (t)$---at each time step over the whole simulation domain. As has been mentioned, a conventional DNN is no longer used to build the ROM in this test case. FNO and DeepONet are able to learn the operator of the PDEs and can be trained to predict the whole time series.

To train the DeepONet-based ROM, the constant input vector is given to the branch net, and the 200-time-step vector is given to the trunk net to predict the two QoIs. As for FNO, the input layer contains the inputs that remain constant throughout the entire simulation. Note that in a future work, we will improve the ROMs by training them on time-dependent inputs as well. But for the present, they are treated as constant throughout the simulations. In both OL algorithms, the output layers predict the $V_{\text{bead}} (t)$ and $T_{\text{mp,max}} (t)$ at each of the time steps. To accurately capture the details of the time-dependent QoIs, we used 50 modes in the FNO model, as time-dependent QoIs are more challenging to fit than the scalar QoIs covered in Section \ref{section:results-maximum-QoIs}. For DeepONet, we used three hidden layers and 130 neurons for both the branch and trunk nets. These and other hyperparameter values (e.g., the learning rate) were optimized in a manner similar to the one presented in Section \ref{section:results-hyperparameter}. However, the exact details of this optimization are not provided here.

\begin{figure}[!htb]
	\centering
	\captionsetup{justification=centering}
	\includegraphics[width=1.0\textwidth] 
        {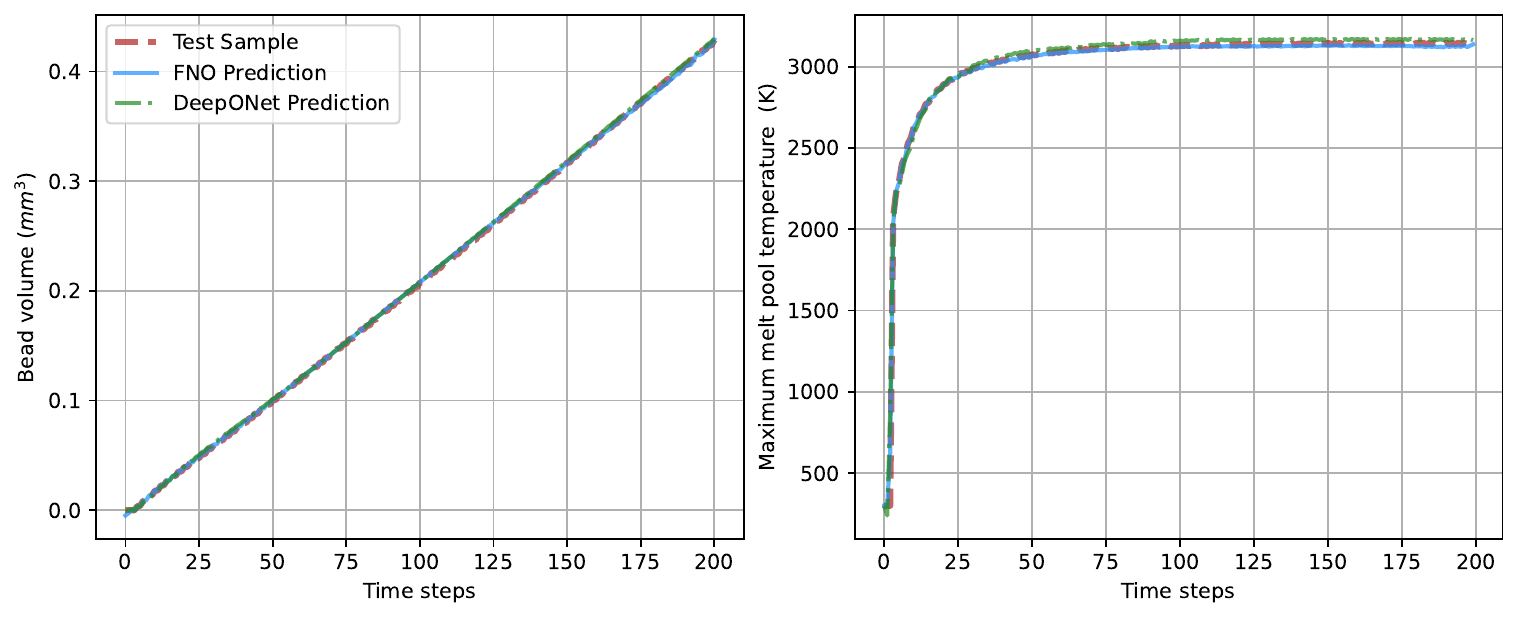}
	\caption[]{Comparing DeepONet and FNO's predictions of $V_{\text{bead}} (t)$ and $T_{\text{mp,max}} (t)$ against the test sample.}
	\label{fig:ROMs-time-pred-vol-temp}
\end{figure}

Figure \ref{fig:ROMs-time-pred-vol-temp} illustrates, for a single test sample, a comparison of the two time-dependent QoIs predicted by DeepONet and FNO against the physical model simulation. Very good agreement between the ROM predictions and the actual test sample values were observed, with similar agreement being found in almost all 47 test samples.

\begin{figure}[!htb]
	\centering
	\captionsetup{justification=centering}
	\includegraphics[width=1.0\textwidth] 
        {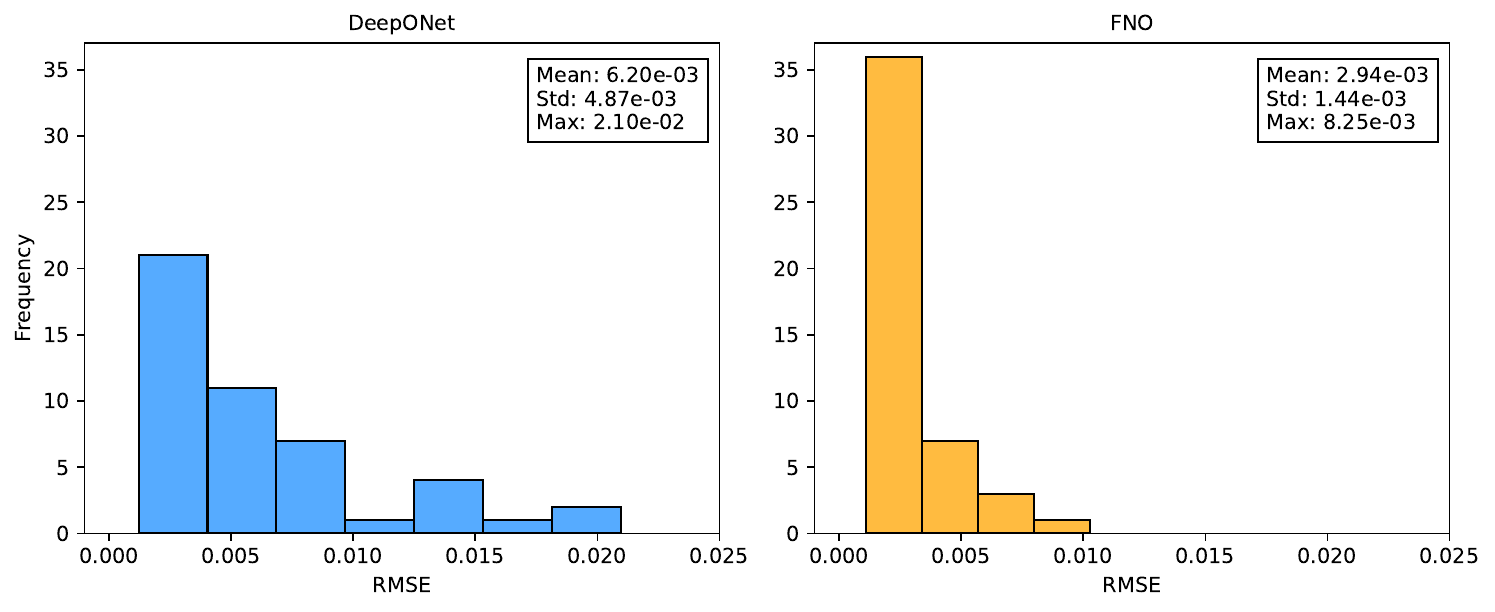}
	\caption[]{RMSE distribution for the ROM predictions of $V_{\text{bead}} (t)$.}
	\label{fig:ROMs-time-pred-RMSE-bead-volume}
\end{figure}

\begin{figure}[!htb]
	\centering
	\captionsetup{justification=centering}
	\includegraphics[width=1.0\textwidth] 
        {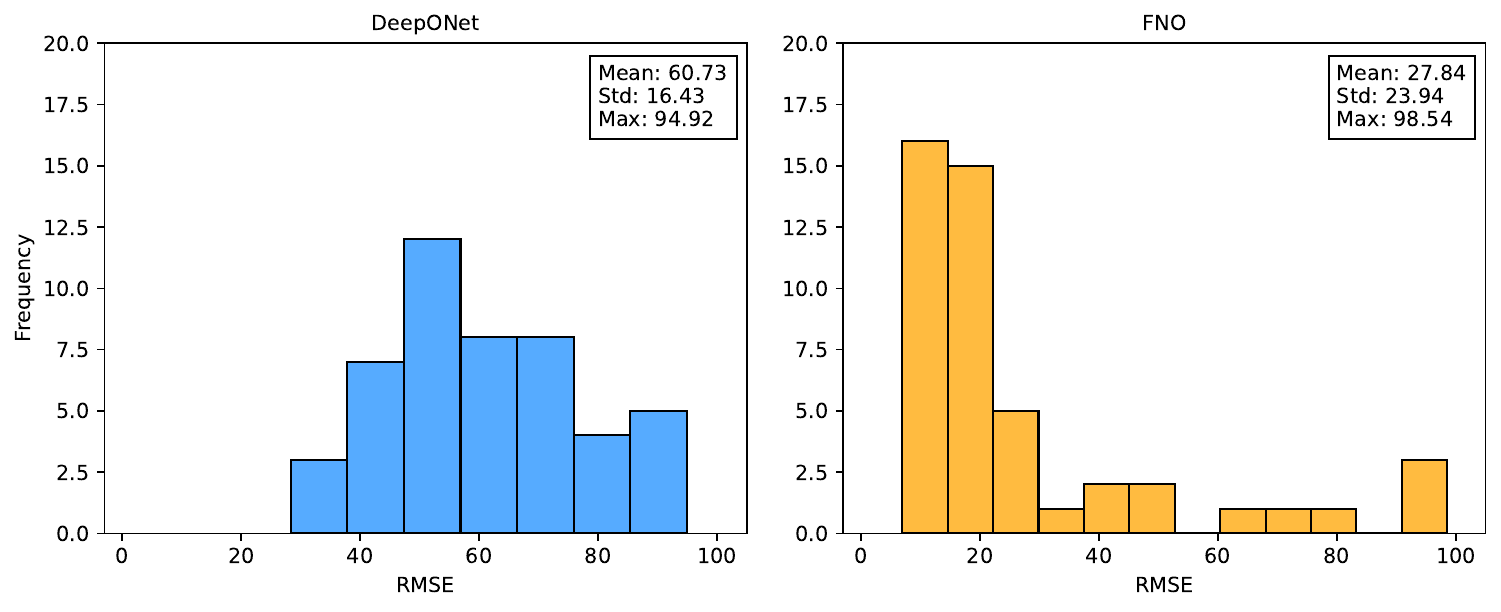}
	\caption[]{RMSE distribution for the ROM predictions of $T_{\text{mp,max}} (t)$.}
	\label{fig:ROMs-time-pred-RMSE-max-temp}
\end{figure}

\begin{figure}[!htb]
	\centering
	\captionsetup{justification=centering}
	\includegraphics[width=1.0\textwidth] 
        {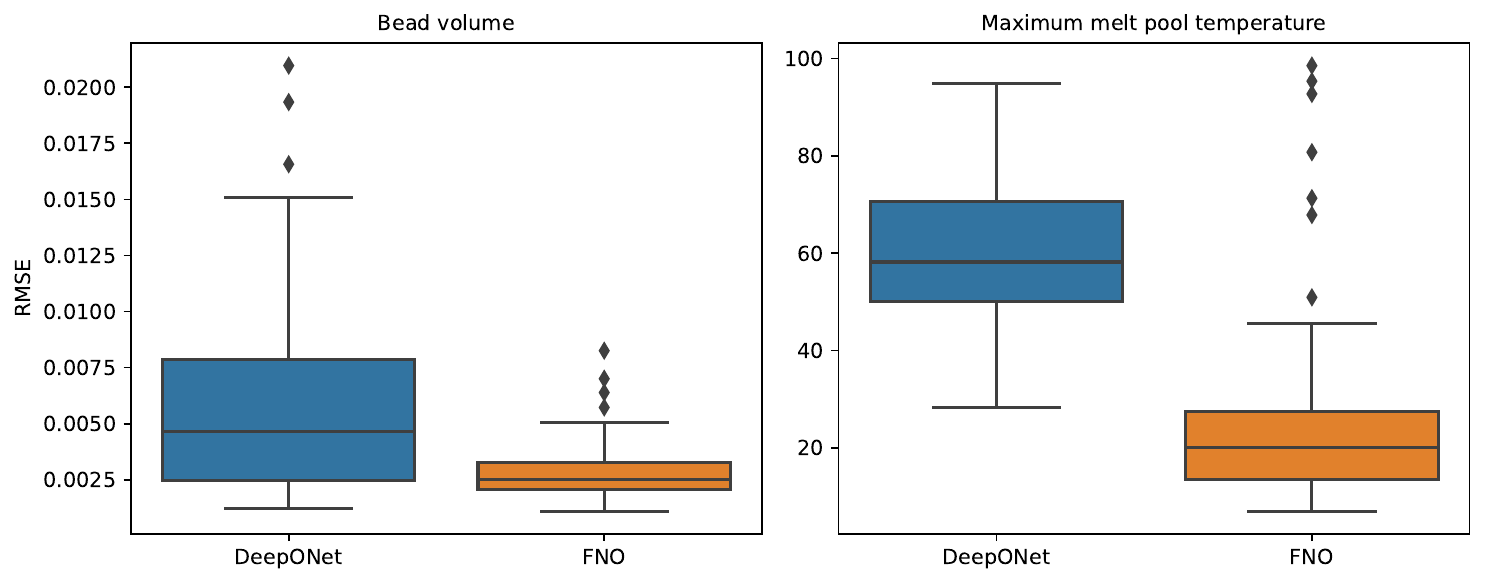}
	\caption[]{Box plots of the RMSEs for the time series 
         predictions generated by DeepONet/FNO.}
	\label{fig:ROMs-time-pred-RMSE-box-plot}
\end{figure}

Figures \ref{fig:ROMs-time-pred-RMSE-bead-volume} and \ref{fig:ROMs-time-pred-RMSE-max-temp} illustrate the RMSE distributions for all the test samples in light of both QoIs. The RMSEs of the FNO model for $V_{\text{bead}} (t)$ show the majority of test samples to be concentrated around a small error value, whereas the DeepONet model exhibits larger, more widely distributed errors. For $T_{\text{mp,max}} (t)$, the FNO model has a smaller mean error than DeepONet. However, its variance is larger than that of DeepONet. By examining the box plot shown in Figure \ref{fig:ROMs-time-pred-RMSE-box-plot}, the average RMSE of the FNO model is seen to be smaller than that of DeepONet for both time-dependent QoIs. The RMSE variance seen in the FNO model is smaller as well.

\section{Conclusions}
\label{section:Conclusions}

The present work employed OL as a ROM approach for the MOOSE-based AM model, enabling preductions of both the bead volume and maximum melt pool temperature. The goal was to construct a fast and accurate ROM of this MOOSE-based AM model---one that can be used in a future AI-based process control and optimization work. We examined two test cases one that considered scalar QoIs and one that considered time-dependent QoIs. In the first case, we focused on predicting the maximum values for the bead volume and melt pool temperature over the whole simulation. For this analysis, we utilized FNO and DeepONet, as they can learn a family of equations, then compared their results against those derived from a conventional DNN-based ROM. Our analysis demonstrated both OL models to be more accurate ROMs than the conventional DNN. DeepONet and FNO showed comparable performance in predicting scalar QoIs, and can effectively substitute for the expensive AM model when applied to multi-query tasks that involve repeated runs of the computational model. While the conventional DNN model was faster to train than FNO and DeepONet as a result of its much simpler structure, it was less accurate and could not generalize as effectively as OL models when applied to new datasets A global sensitivity study showed both QoIs to be most sensitive to the scaling factor input parameter, and there is negligible interaction among the computational model input parameters. In the second test case, which focused on time-dependent QoIs, the ROMs were trained to predict the time series QoIs over the whole simulation. Both OL models showed good performance in predicting time-dependent QoIs. FNO generally showed a smaller generalization error than DeepONet, but also a larger dispersed error distribution due to some outlier predictions. Both FNO and DeepONet can be trained to predict time series without the need for dimensionality reduction, unlike when using a conventional DNN to train a ROM for time series responses. In general, OL methods outperform DNN at building ROMs for complex processes such as AM, and can very accurately predict both scalar values and the time series of the QoIs. In future work, we will improve the OL-based ROMs such that they can take time-dependent inputs as well. Furthermore, the trained ROMs will be integrated into a deep reinforcement-learning-based process control and optimization framework to search for the optimal design variables and adaptive system settings for enhancing the properties of the final AM product.

\section*{Acknowledgement}

This work was supported through the INL Laboratory Directed Research \& Development (LDRD) Program under DOE Idaho Operations Office contract no. DE-AC07-05ID14517. This research made use of Idaho National Laboratory computing resources that are supported by the Office of Nuclear Energy of the U.S. Department of Energy and the Nuclear Science User Facilities under contract no. DE-AC07-05ID14517.


\bibliography{./bibliography.bib}

\end{document}